\documentclass[10pt,twocolumn,letterpaper]{article}

\usepackage{cvpr}
\usepackage{times}
\usepackage{epsfig}
\usepackage{graphicx}
\usepackage{amsmath}
\usepackage{amssymb}

\usepackage[pagebackref=true,breaklinks=true,letterpaper=true,colorlinks,bookmarks=false]{hyperref}

\cvprfinalcopy 

\def\cvprPaperID{2973} 

\ifcvprfinal\fi
\begin{document}

\title{WAYLA - Generating Images from Eye Movements}

\author{Bingqing Yu\\
Centre for Intelligent Machines, McGill University\\
{\tt\small bingqing.yu@mail.mcgill.ca}
\and
James J. Clark\\
Centre for Intelligent Machines, McGill University\\
{\tt\small clark@cim.mcgill.ca}
}
\maketitle

\begin{abstract}
We present a method for reconstructing images viewed by observers based only on their eye movements. By exploring the relationships between gaze patterns and image stimuli, the ``What Are You Looking At?" (WAYLA) system learns to synthesize photo-realistic images that are similar to the original pictures being viewed. The WAYLA approach is based on the Conditional Generative Adversarial Network (Conditional GAN) image-to-image translation technique of Isola \etal \cite{isola2016image}. We consider two specific applications - the first, of reconstructing newspaper images from gaze heat maps, and the second, of detailed reconstruction of images containing only text. The newspaper image reconstruction process is divided into two image-to-image translation operations, the first mapping gaze heat maps into image segmentations, and the second mapping the generated segmentation into a newspaper image. We validate the performance of our approach using various evaluation metrics, along with human visual inspection. All results confirm the ability of our network to perform image generation tasks using eye tracking data.
\end{abstract}

\section{Introduction}

Image generation has always been one of the primary topics in the field of computer vision. Due to various limitations such as the lack of source image information, it often happens that the available image datasets are either insufficient in quantity or defective in quality. Therefore, a variety of image synthesis methods have been developed to generate images that are more useful and valuable for subsequent image processing tasks.

In past years, various linear and nonlinear methods have been explored to improve image quality via interpolation \cite{hou1978cubic,unser1999splines,zhang2006edge}. Recently, researchers have focused on using super-resolution approaches to generating more detailed images \cite{farsiu2004fast,liu2016learning}.

One of the shortcomings of the aforementioned image enhancement approaches is that, for most of them, they only consider the problem of generating an image with higher resolution using a given low-resolution image. However, in many instances, one doesn't even have a lower resolution image. How can we generate an image in this case? Obviously, some additional information is needed. One approach that is quite popular recently is to train a network to generate an image based on a verbal description. In this paper we look at another situation, wherein we have someone looking at an image, and want to generate an image similar, if not identical, to the one that is being viewed. But we assume that the only information we have available is the viewers' gaze patterns or eye movement trajectories.

It is not immediately obvious whether there is enough information in the eye tracking data to allow reconstruction of what is being viewed. But, there is a well-examined link between image content and gaze patterns. It has been stated by many studies related to visual attention that viewers scan the interesting locations of a given image by controlling their attention, and the trajectory of the attention can be explored by tracking the eye movements of the viewers. Examples in the literature include the Clark and Ferrier computational model of saliency which was used in a robotic system to demonstrate the relationships that exist between image saliency and eye movements \cite{clark1988modal} and the Itti \etal visual attention model \cite{itti1998model}, which also showed the ability of the eye fixation data to represent viewing behaviors and image salience characteristics. In the context of attention tracking during text reading, researchers have carried various studies and developed numerous theories and models in order to describe how the eyes move while reading. For example, O'Regan \etal \cite{o1987eye} demonstrated that for a single word recognition task, there exists an initial fixation location which minimizes the probability of refixation and total fixation duration on a word. In the context of continuous reading, this was extended to the notion of Preferred Viewing Location, where the fixation locations depend on the words being read \cite{mcconkie1988eye}. McConkie and co-workers have discovered that there is an optimal fixation position in sentence reading, which is slightly to the left of a word center. Its duration can be influenced by various factors such as the complexity of a word. Moreover, it has been found that the fixation durations are the longest if the target is at the word center \cite{vitu2001fixation}. Grammatical factors also have an impact: functional words such as prepositions often have a higher skipping rate \cite{wotschack2009eye}. In sum, reading behaviors and fixation characteristics are strongly influenced by the target locations and various lexical variables of the reading material. This suggests that the particular shape and structure of eye movement trajectories constrain, to some unknown extent, the text being read. Similar conclusions can be drawn with respect to general image viewing - that the eye trajectory information constrain to some extent the contents of the image being viewed.

\subsection{Proposed Method}

To investigate the problem of generating images from eye movements, we propose to use deep learning and create a neural network which takes in eye fixation data or gaze heat maps and generates as output images similar to the original scenes that are displayed to the viewers. We call our approach "WAYLA", which stands for "What Are You Looking At?". The feasibility of this approach is enhanced by the presence of numerous datasets of images with corresponding eye tracking annotations that have been published in recent years \cite{winkler2013overview}. 

Although it is possible to construct an image generation model from scratch, we benefit from the state-of-the-art deep learning models to perform our eye movements to image prediction task. We propose to use a network based on the Conditional Generative Adversarial Network, the details of which will be presented in Section~\ref{architectureandtraining} \cite{mirza2014conditional}. Instead of using solely the idea of the traditional GAN network \cite{goodfellow2014generative} to generate images by minimizing the Euclidean distance between output images and ground truth stimuli, we use the architecture of the Conditional GAN. In this way, our network can learn to generate a synthetic image by conditioning on a corresponding eye fixation heat map. As a result, the goal becomes to minimize the difference between a patch combining the eye fixation heat map with the generated image and a patch combining the eye fixation heat map with the ground truth image. The architecture of the proposed WAYLA network uses the image-to-image translation model presented by Isola \etal \cite{isola2016image} as a pre-built model. This study focuses on its specific application, which is transforming an image from one version to another; however, we focus on exploring it as a solution to the eye-fixation-data-to-image generation problem. We mainly focused on applying our model to two eye tracking datasets \cite{vil2013algorithm,cop2017presenting}. One contains various scanned images of newspapers and magazines with associated gaze heat maps, and the other contains eye tracking data obtained during text reading. Our network takes the eye fixation heat maps provided by these datasets as input and produces two sorts of output images. The first type of output is a simplified semantic segmentation, while the second type of output is a detailed photorealistic image. In particular, our network is able to synthesize a segmented version of newspaper appearances. This simplified version of realistic newspapers is consistent with the eye-tracking data and filled with semantically labeled regions representing picture and text. Taking a step further, our network is also effective at then generating text-embedded images that are at a higher detail level. In this way, it is possible to obtain a more concrete representation of what people are looking at during reading. The dataset provided by Vilkin \etal \cite{vil2013algorithm} contains numerous highly detailed newspaper images and their corresponding segmented versions. Since this database did not have any associated eye track data, we used a state-of-the-art saliency model, MLNET \cite{cornia2016deep}, to generate all the eye fixation heat maps for the ground truth newspaper images. These heat maps are used to model where people may look while viewing these newspapers, and serve as the eye fixation data that is used as input for our network during training. We also applied our WAYLA approach to another dataset, named GECO, the Ghent Eye-Tracking Corpus \cite{cop2017presenting}. The GECO dataset contains eye fixation data collected during reading of text on pages of novels displayed on a computer screen. 

We employ a conditional generative adversarial network in our implementation. Where the eye fixation data is used to condition the operation of the generator. The discriminator serves to distinguish between ``fake'' segmentations or images and ``real" ones. Therefore, in the case of training on ``real'' examples, the discriminator input consists of an image which is a combination of the eye fixation data and the ground truth images. The ground truth images are newspaper images at various detail levels for the Vilkin \etal dataset and text-embedded images for the GECO dataset.  However, in the case of ``fake" examples, the combination is a concatenation of the eye fixation data with the output images produced by the generator. In this way, by training the network on the examples described above, it learns to generate synthetic images that are similar to the ground truth images.

\section{Architecture and Training}
\label{architectureandtraining}

\subsection{Data Preparation}

\begin{figure}[t]
\begin{center}
\fbox{
   \includegraphics[width=0.9\linewidth,height=3cm]{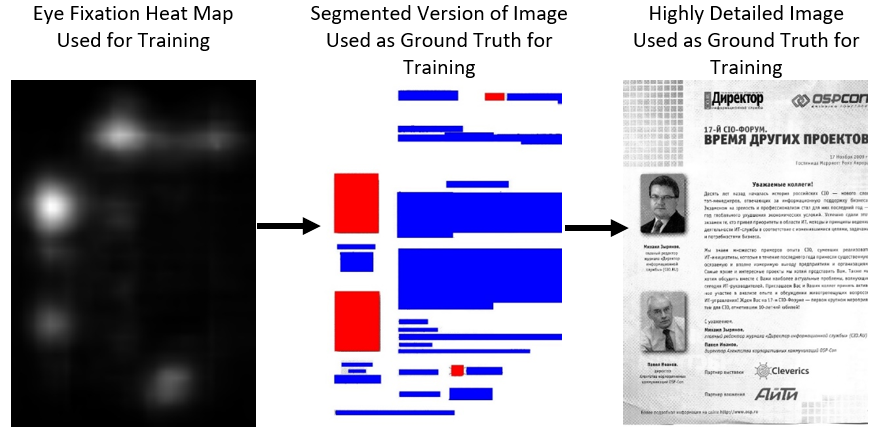}}
\end{center}
   \caption{Illustration of our image generation pipeline. Our approach formulates the problem of generating detailed newspaper images from eye fixation data as a two-phase process. The network needs to be fed with the eye fixation heat maps as input and trained with the segmented images as ground truth. Then we can utilize the segmented images, along with the eye fixation data, to train the network with the highly detailed images as ground truth.}
\label{fig:pipelinegt}
\end{figure}

For training our network to learn a mapping from eye movement data to newspaper images provided by the Vilkin \etal dataset, we needed to produce eye fixation data and feed them as input for our model. For this purpose we used the state-of-the-art MLNET model \cite{cornia2016deep} to build the eye fixation heat map dataset. To obtain the MLNET saliency predictions for the newspaper dataset, we fed the MLNET network with the original scanned newspapers used by Vilkin \etal, and used this saliency model as the corresponding salience heat map for each input stimulus. This allows us to get fixation heat maps that are strongly in agreement with the real gaze locations. In this way, by using the generated eye fixation data and the image datasets provided by the Vilkin dataset, we can train our model to output newspaper images at various detail levels. As shown in Figure~\ref{fig:pipelinegt}, we broke the end-to-end image generation process into two phases. The first phase has the goal of generating semantic segmentations of the newspaper images, while the second phase is used to generate the detailed newspaper images from the segmentations.

\begin{figure}[t]
\begin{center}
\fbox{
   \includegraphics[width=0.9\linewidth]{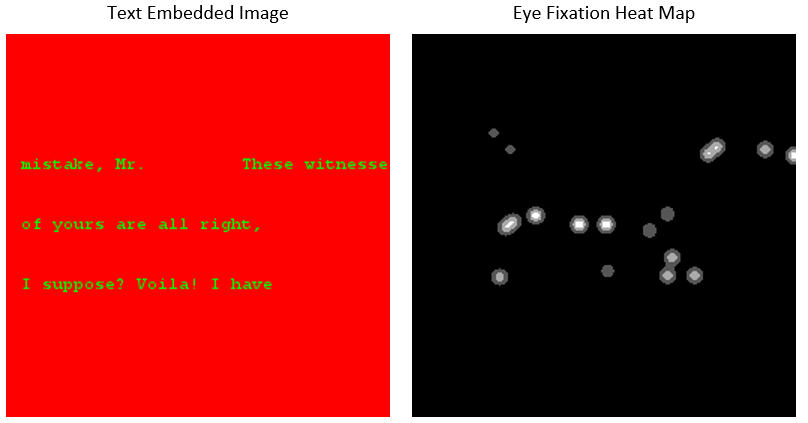}}
\end{center}
   \caption{(Left) A ground truth text-embedded image generated from the GECO dataset. (Right) The corresponding eye fixation heat map generated from the fixation data of the GECO dataset.}
\label{fig:eyetotext_textgen_gt}
\end{figure}

We also investigated the effectiveness of our model on reconstructing the text images used in obtaining the GECO eye movement dataset. This dataset, which contains records about the eye fixation positions and durations of each individual for each reading session, enabled us to directly use their eye fixation data as input to our model. Therefore, as shown in Figure~\ref{fig:eyetotext_textgen_gt}, we were able to generate the eye fixation heat maps that correspond to different parts of the novel that were read by participants. When generating grayscale eye fixation heat maps for the GECO dataset, for each observer and for each time a fixation is made on a specific position of a specific word, we place a bright point in the grayscale heat map at a position that corresponds to its recorded fixation location. The brightness of that point is modulated with the recorded percentage of time spent on that specific location out of the total trial time completed by an observer. The maximum value for this recorded percentage value is 0.17 percent, so all fixation points with fixation duration percentage less than this value will be represented with a less bright point in our heat maps. The maximum pixel value for the synthetic heat map is 255, corresponding to a fixation point that has a duration value of 0.17 percent. If a fixation occupies 0.017 percent of the total trial time, its pixel value becomes 25.5. It is possible that several fixations are made within one word, in this case, distinct bright points corresponding to distinct fixation positions will be added into the heat map. However, for the fixation points belonging to the same word, we chose to modulate the brightness of all these fixation points using the total percentage of trial time for that specific word under the assumption that the global duration value is more useful for estimating the importance of a word as compared to other words in the reading material.

As for the ground truth images that serve as targets for training our network, no sophisticated data preparation process was undertaken. The ground truth segmented newspaper images and ground truth detailed newspaper images are simply taken from the dataset provided by Vilkin \etal.  Regarding the GECO dataset, we chose to break the reading material into parts for generating RGB images containing printed text. Each text image is a (256 X 256) size RGB image, with its red channel encoding a constant background and its green channel encoding text content. The blue channel is set to zero everywhere. It was found experimentally that this 3-channel arrangement provided better training stability, reduces the possibility of divergence and allows faster convergence, as compared with using a single channel containing only the text content. Each image contains 15 words arranged into 3 rows and each row contains 5 words one following another.  For generating our eye fixation heat maps for the GECO dataset, the locations of all saliency points are adapted to the locations of the generated text-embedded images. We ensured that the correspondence is in agreement with the data provided by the GECO dataset.

\subsection{Network Training}

\begin{figure}[t]
\begin{center}
\fbox{
   \includegraphics[width=0.9\linewidth]{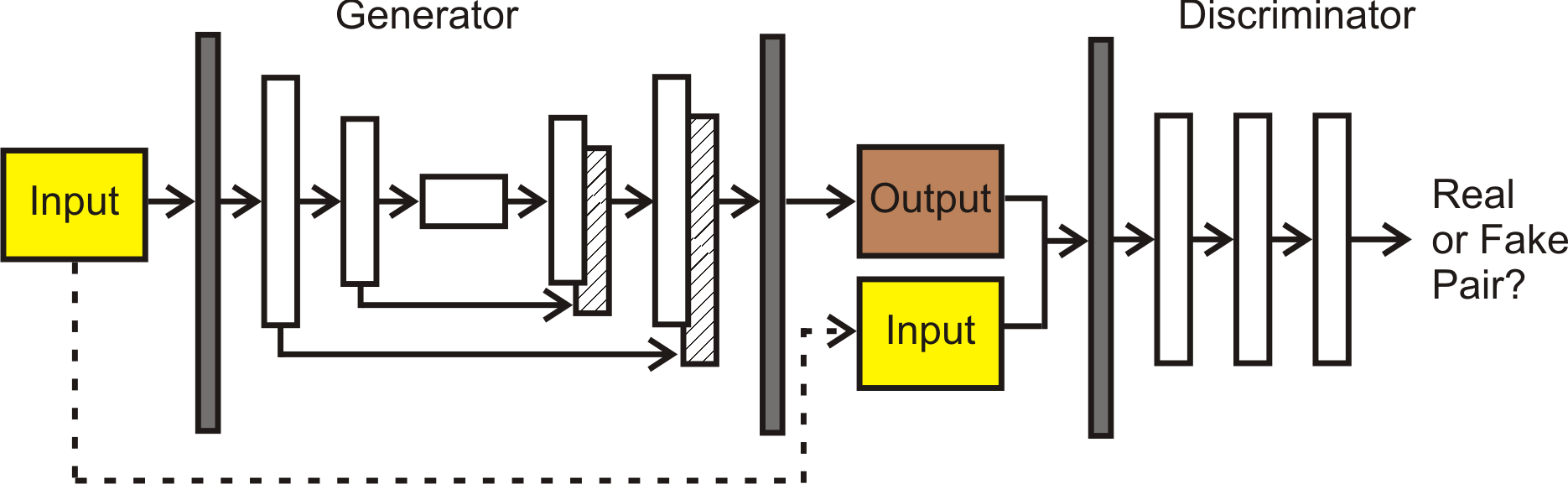}}
\end{center}
   \caption{Illustration of the Conditional GAN architecture used for WAYLA.}
\label{fig:congan}
\end{figure}

As stated earlier, the relationships between eye movements and viewing material have been examined in numerous studies, giving us confidence that we can develop an image synthesis method that is able to synthesize images from eye track data that are similar to some extent to the images actually being viewed. In recent years, generative adversarial networks and their extensions led to advances in the field of photo-realistic image synthesis \cite{goodfellow2014generative,radford2015unsupervised,mirza2014conditional,isola2016image}. Given enough training samples, they can produce photo-realistic images that are very close to the ground truth pictures. Therefore, we chose to build our system based on the architecture of Conditional GAN, which has proven to be an efficient way to do image-to-image translation for various computer vision tasks such as image segmentation and grayscale-to-RGB image translation \cite{isola2016image}. The overall architecture based on which the WAYLA system is built is illustrated in Figure~\ref{fig:congan}.

The input data was the gaze heat maps obtained from the aforementioned data preparation steps, and we fed these heat maps to the input layer of our neural network using the architecture of Conditional GAN described in the work of Isola P. \etal \cite{isola2016image}. The modifications that we made to the published conditional GAN structure are presented in the remaining part of the section.

\subsubsection{Individual Training of Two Phases for Newspaper Generation}
\label{individual}

We proposed WAYLA as a model to generate image content based on eye fixation data. Since the Vilkin \etal dataset provided us with both segmented and detailed newspaper images, we formulated the image generation task as a two-phase process. The first phase consists in training the network to do eye-movement-data-to-segmented-newspaper-image synthesis. The second phase consists in training the network to generate newspaper images with a higher level of detail from the image segmentation. In this section, we will present how we generated newspaper images by training the WAYLA model separately and independently for the two phases. In the next section, we will present a slightly different approach that joins the two training phases to yield an End-to-End implementation for generating newspaper images.

\begin{figure}[t]
\begin{center}
\fbox{
   \includegraphics[width=0.9\linewidth, height= 2.5cm]{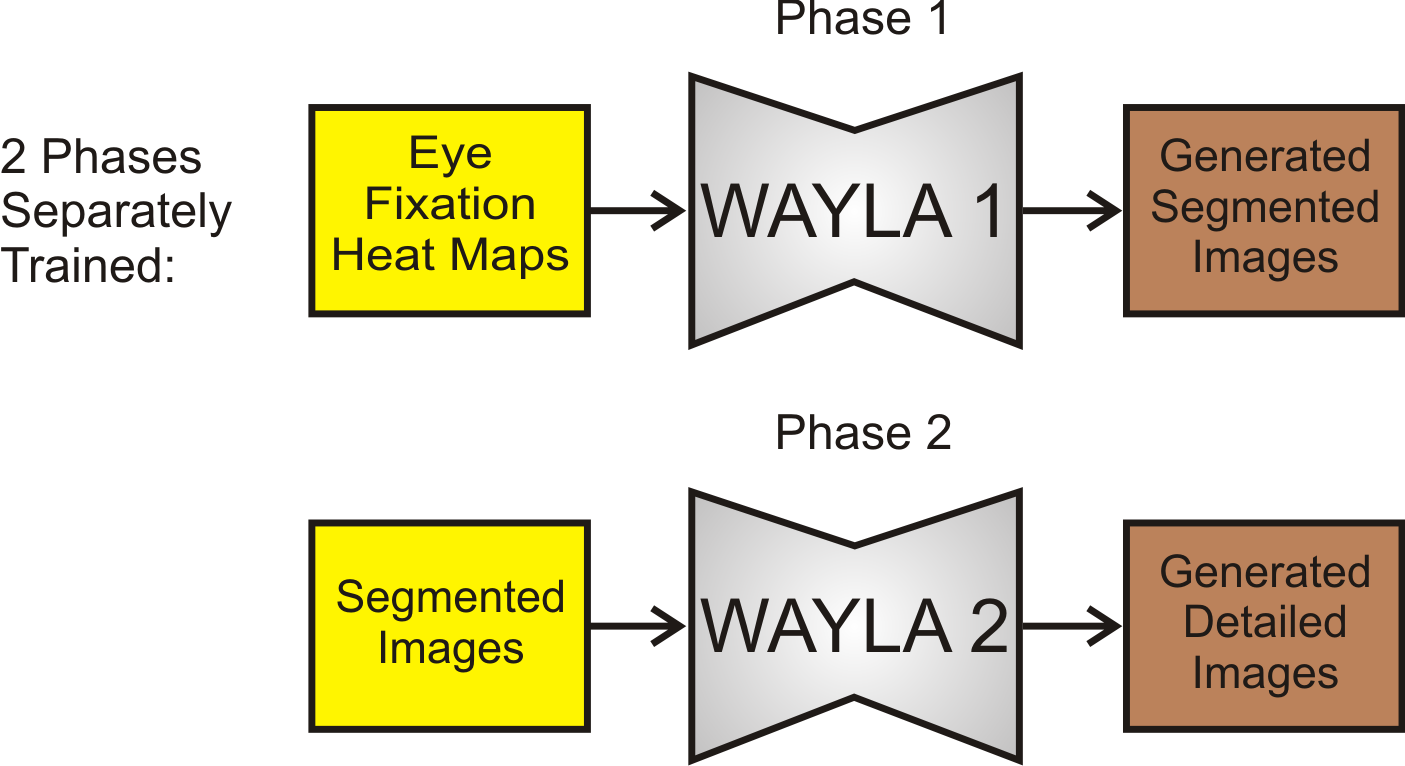}}
\end{center}
   \caption{Illustration of the implementation which trains the network separately for generating segmented images and detailed images. The top part of the figure shows the training process of the first phase. Our model takes the eye fixation data as input and is trained with the segmented newspaper images as ground truth. The bottom part of the figures shows the training process of the second phase. Our model takes the segmented images from the Vilkin \etal dataset as input and is trained with the detailed newspaper images as ground truth. }
\label{fig:2phases}
\end{figure}

When training our network for the first phase, the generator is fed with the eye fixation heat maps that we produced using MLNET. During training, the generator is optimized to produce outputs that are as similar as possible to the ground truth segmented newspaper images. The discriminator is fed with an image patch which concatenates the input eye fixation heat maps with the generated images produced from the generator. When receiving this kind of patches, the discriminator is trained to recognize them as ``fake" images. In the ``real" image case, the discriminator receives patches which concatenate the eye fixation heat maps with the ground truth segmented newspaper images. An illustration of this first phase training is shown in the upper part of Figure~\ref{fig:2phases}.

For the second phase, we trained our network to synthesize detailed newspaper images based on the segmented newspaper images. The lower part of Figure~\ref{fig:2phases} illustrates the input and output settings of this second phase training. During the second phase, the input layer of the generator is fed with the segmented images provided by the Vilkin \etal dataset. Then, the generator is optimized to produce outputs that are as similar as possible to the ground truth highly detailed newspaper images. In this case, the discriminator is fed with image patches which concatenate the segmented images and the detailed images, and its task is to distinguish synthesized data from ground truth data.

\subsubsection{End-to-End Design of Pipeline for Newspaper Generation}

We also explored the possibility of our network to learn to generate highly detailed newspaper images when it only has the eye fixation heat maps at its disposal, which is the more interesting case.

To achieve this goal, we applied an end-to-end training process to our system. We started by feeding the input layer of generator with the eye fixation heat maps and training the system to generate segmented images by using the segmented newspapers from the Vilkin \etal dataset as ground truth. At this stage, the operation is exactly the same as the first phase training presented in Section~\ref{individual}. However, the difference between the implementation mentioned in Section~\ref{individual} and this end-to-end implementation is indicated by the input and output settings that we used for the remaining part of the training process. After finishing training our system to produce segmented images, we re-initialized the system and fed the input layer of the generator with a new kind of input, different from the segmented images that we used as input for the second phase training presented in Section~\ref{individual}. At this time, the segmented images generated by the generator that we just trained previously are concatenated with the eye fixation heat maps to form a new set of input RGB images, which are then fed into our re-initialized system. Now, the generator takes these inputs and is optimized to output images that are as similar as possible to the ground truth highly detailed newspaper images provided by the Vilkin \etal dataset. It is worth mentioning that the concatenation is done in such a way that the new red channel is formed by adding the pixel values of the eye fixation heat maps to the pixels values of the red channel of the generated segmented images. The new blue channel is formed by taking the pixel values of the blue channel of the generated segmented images. The new green channel is formed by setting all the values to 0, except for the locations where all the three channels of the generated segmented are equal to 255, in this case, the green channel pixel remains 255 to form a white color along with the other two channels. The design of the discriminator is the same as all the aforementioned designs. It receives image patches and distinguishes whether they belong to ``true image pairs" or ``fake image pairs".

\begin{figure}[t]
\begin{center}
\fbox{
   \includegraphics[width=0.9\linewidth]{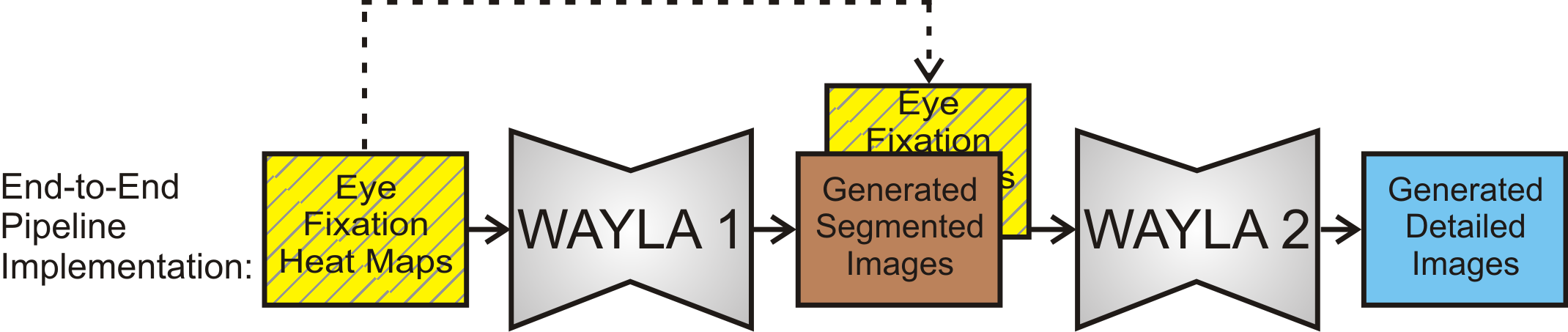}}
\end{center}
   \caption{Illustration of the end-to-end implementation. Initially our model takes the eye fixation data as input and is trained with the segmented newspaper images as ground truth. Then, the model takes as input image patches which concatenate the eye fixation data with the generated segmented images, and it is trained with the highly detailed newspaper images as ground truth. }
\label{fig:endtoend}
\end{figure}

In this way, we explored the feasibility of generating highly detailed newspaper images when only the eye fixation heat maps are available. An illustration of the end-to-end design is shown in Figure~\ref{fig:endtoend}. 

\subsubsection{Training on the GECO Dataset}

We also applied WAYLA on the GECO dataset in order to investigate the effectiveness of our model to generate text-only images based on eye fixation data. The generator has as input the eye fixation heat maps that we created from the GECO dataset. While the generator is trained to produce text-like images and has the ground truth text embedded images as target, the discriminator is trained to distinguish the generator's outputs as ``fake" images. In the ``fake image" case, the discriminator receives as input a combination of the eye fixation heat maps and the generator's outputs; however, in the ``real image" case, the discriminator receives the eye fixation heat maps concatenated with the ground truth text embedded images as input.

The loss functions used for our network are presented in the following, and it is applied on all training phases and all datasets involved in our study.

The discriminator, whose task is to classify between real and fake pairs, uses the following binary cross entropy loss as its loss function: 

\begin{equation}
L_D = E_{x,y}[log D(x,y) ] + E_{x}([1-log D(x,G(x)) ]
\label{equ:disc}
\end{equation}

In Equation~\ref{equ:disc}, $x$ is the input of the generator, $y$ represents all ground truth images that the generator has as target. As for the generator, since it is stated in \cite{pathak2016context} that mixing the GAN loss with another standard content loss such as Euclidean loss can improve the training of deep neural networks, we chose to use the $L_1$ distance as the additional loss and combine it with the adversarial loss described above to construct the loss function for our generator. The $L_1$ distance represents the difference between the outputs of the generator and the ground truth images. Thus, the overall loss function of the generator is defined as:

\begin{equation}
L_G = L_D + \lambda L_1(G)
\label{equ:gen}
\end{equation}

We set the value of  $\lambda$  to 0.01, and this decision is taken based on our observation from experiments and on the analysis made in \cite{isola2016image}, which states that when the $L_1$ loss is weighted 100 time larger than the GAN loss, there are fewer artifacts produced in the output of the generator. All layers of the network need to be trained from scratch. Weights are randomly initialized using uniform distribution between $-0.05$ to $0.05$.  We always preserve 20 percent of the total samples for testing. The network is trained by updating the generator and the discriminator alternatively. The GAN cross-entropy loss is backpropagated to the discriminator to update its weights. Then, by keeping the discriminator weights constant, we combine the cross-entropy loss with the $L_1$ loss and backpropagate this error to update the generator weights. Minibatches of 2 samples per batch are used for training. The RMSProp optimizer is used to optimize both the generator and the discriminator, with a learning rate of 0.001, a decay rate of 0.9, a momentum of 0 and an $\epsilon$ of $1\times10^{-6}$. Dropout layers and batch normalization are used in our network to accelerate convergence.

\section{Evaluation and Discussion }

\begin{table}
\begin{center}
\begin{tabular}{|l|c|}
\hline
 & SSIM \\
\hline\hline
Segmented newspapers & 0.83 \\
\hline
\shortstack[l]{Detailed newspapers \\ (Individual Training) } & 0.54 \\
\hline
\shortstack[l]{Detailed newspapers \\ (End-to-End) } & 0.53 \\

\hline
\end{tabular}
\end{center}
\caption{SSIM scores obtained by comparing our synthesized images with the ground truth.}
\label{table:ssim}
\end{table}

Among the large body of work done on evaluating the perceptual quality of images, losses such as L1 and L2 distances have been the dominant performance metrics in the domain of computer vision. However, in order to address a drawback of the basic L1 and L2 metrics, which is their inability to indicate structural information carried by images, we chose to use the structural similarity (SSIM) index to compute the perceived similarity between our generated images and our ground truth images. In this way, the evaluation metric may provide a better understanding of the overall quality of generated images, since the similarity scores are not only in terms of pixel values but also in terms of image structure \cite{wang2009mean,boulogne2014scikit}. In Table~\ref{table:ssim}, similarity scores are reported for both the segmented and detailed newspaper image generation. By computing the scores between our generated segmented newspaper images and the ground truth segmented newspapers, we can evaluate the ability of our network to do eye-movement-data-to-segmented-image prediction. By computing the similarity scores between our generated detailed newspaper images and the ground truth detailed newspapers, we can evaluate the ability of our network to do the segmented-image-to-highly-detailed-image transformation.  

As there are no other methods for generating images from eye movement data, to investigate the significance of our contribution to synthesizing newspaper images, we looked at the SSIM scores reported in various papers concerning photo-realistic image generation. In comparing our SSIM scores with other studies, our SSIM scores indicate that our synthesized images are similar to the ground truth images. Mihaela R. \etal stated that with their auto-encoding GAN, they can generate images with a SSIM mean value of 0.62 when compared with ground truth images \cite{rosca2017variational}. In another paper, which uses a GAN structure to implement image super-resolution, it is shown that a generated image with SSIM score of 0.6423 is very similar to the original image to human perception \cite{ledig2016photo}. In addition, when Zhou \etal introduced the concept of SSIM, they presented an image disturbed with Gaussian noise with a SSIM score of 0.30. The noisy version still preserves most of the texture and structure characteristics of the original image, so it is still understandable and recognized by human inspection. The SSIM score obtained from our generated detailed newspaper images using the end-to-end implementation is situated slightly below the other SSIM scores presented in Table~\ref{table:ssim}; however, it is understandable that the end-to-end implementation, which trains the network to output highly detailed images without using the ground truth segmented images, can only synthesize images of limited quality compared to the case when the two phases are trained individually. By visually inspecting the qualitative results and comparing them to the ground truth, it can be confirmed the effectiveness of our model at synthesizing real images with eye data. 

\begin{figure}[t]
\begin{center}
\fbox{
   \includegraphics[width=0.9\linewidth]{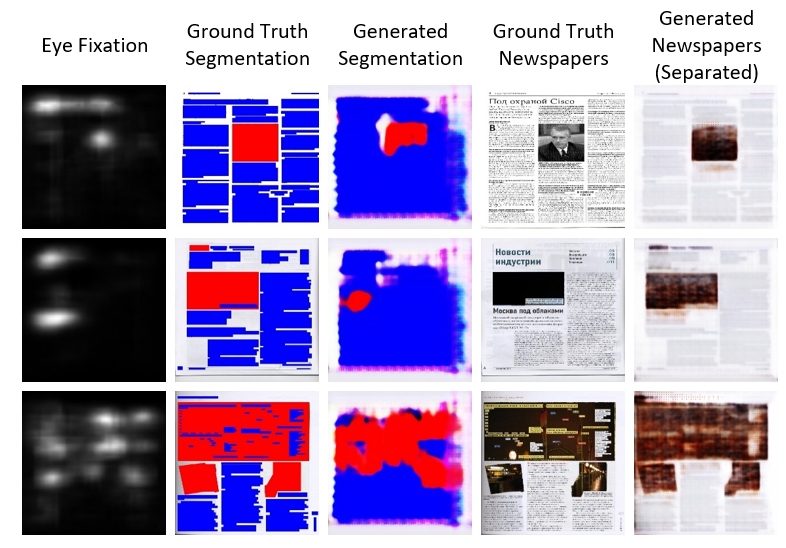}}
\end{center}
   \caption{Example results of WAYLA obtained by training the two phases separately and independently. The qualitative results of WAYLA on the testing set are compared to ground truth.}
\label{fig:pipeline_separate}
\end{figure}

\begin{figure}[t]
\begin{center}
\fbox{
   \includegraphics[width=0.9\linewidth]{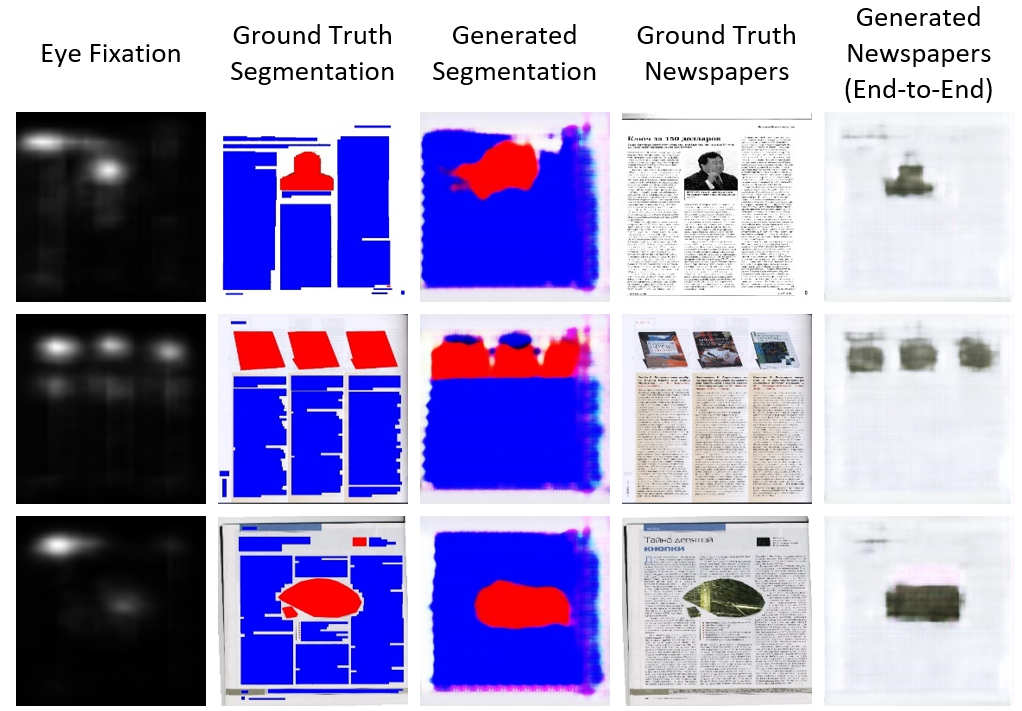}}
\end{center}
   \caption{Example results of WAYLA obtained by implementing the end-to-end training pipeline. The qualitative results of WAYLA on the testing set are compared to ground truth.}
\label{fig:pipeline_end}
\end{figure}

Figures~\ref{fig:pipeline_separate} and ~\ref{fig:pipeline_end} allow a qualitatively evaluation of the performance of our image generation pipeline. It can be observed that the generated segmented images produce text/picture patterns that visually appear close to the ground truth. It is also worth noticing that when WAYLA is applied on a specific dataset, which in our case is the newspaper dataset, and trained for generating segmented images, the network is able to extract meaningful information from eye movement data in such a way that the structure of the generated images fits the specific dataset very well. Taking the first row of Figure~\ref{fig:pipeline_separate} as an example, despite the fact that there are numerous eye fixation points in the eye fixation image, in the generated segmented image, only part of the eye fixation locations are converted into picture areas. The rest of the eye fixation locations are successfully identified as text areas in the generated result. By visually inspecting the detailed images generated by WAYLA, the image patterns of all the generated images can be easily related to the image patterns of the ground truth scanned newspapers presented in the original dataset.


\begin{figure}[t]
\begin{center}
\fbox{
   \includegraphics[width=0.9\linewidth]{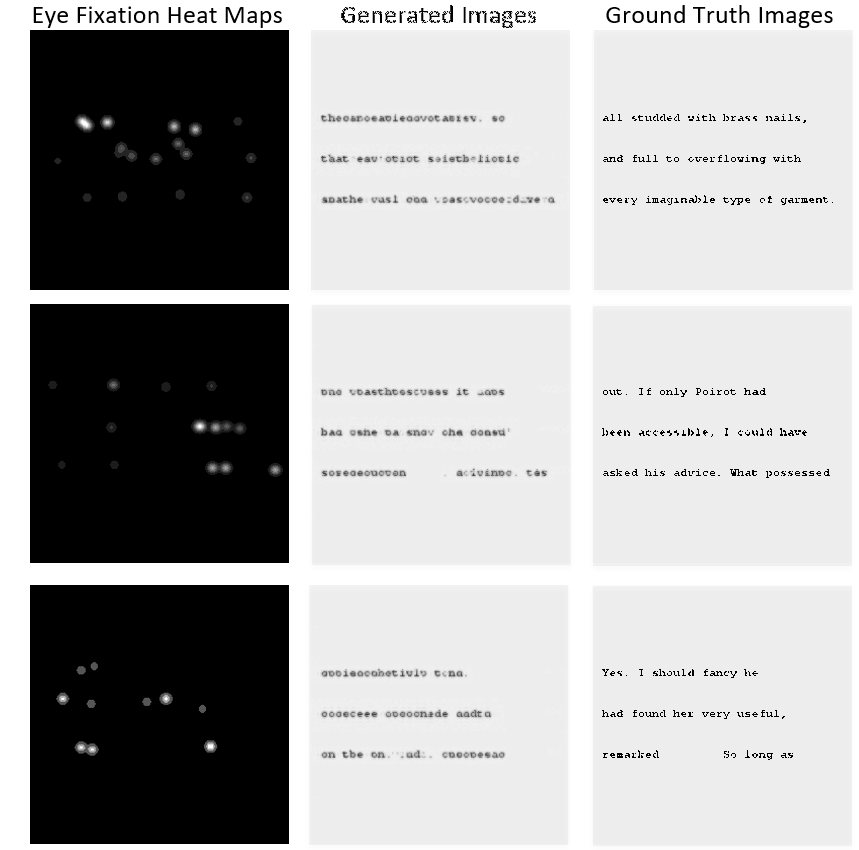}}
\end{center}
   \caption{Qualitative results obtained by training WAYLA to generate text-embedded images using the GECO dataset.}
\label{fig:exampleresults}
\end{figure}

\begin{figure}[t]
\begin{center}
\fbox{
   \includegraphics[width=0.9\linewidth,height= 3cm]{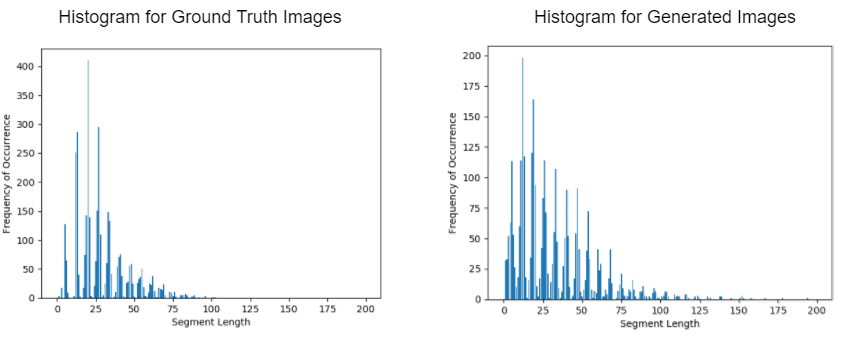}}
\end{center}
   \caption{Comparison between the word length distribution obtained from the ground truth images versus the word length distribution obtained from our generated images. The horizontal axis is the word segment length with unit (pixel). The vertical axis indicates the frequency of occurrence for different word segment length values.}
\label{fig:combo_histo}
\end{figure}

\begin{figure}[t]
\begin{center}
\fbox{
   \includegraphics[width=0.9\linewidth]{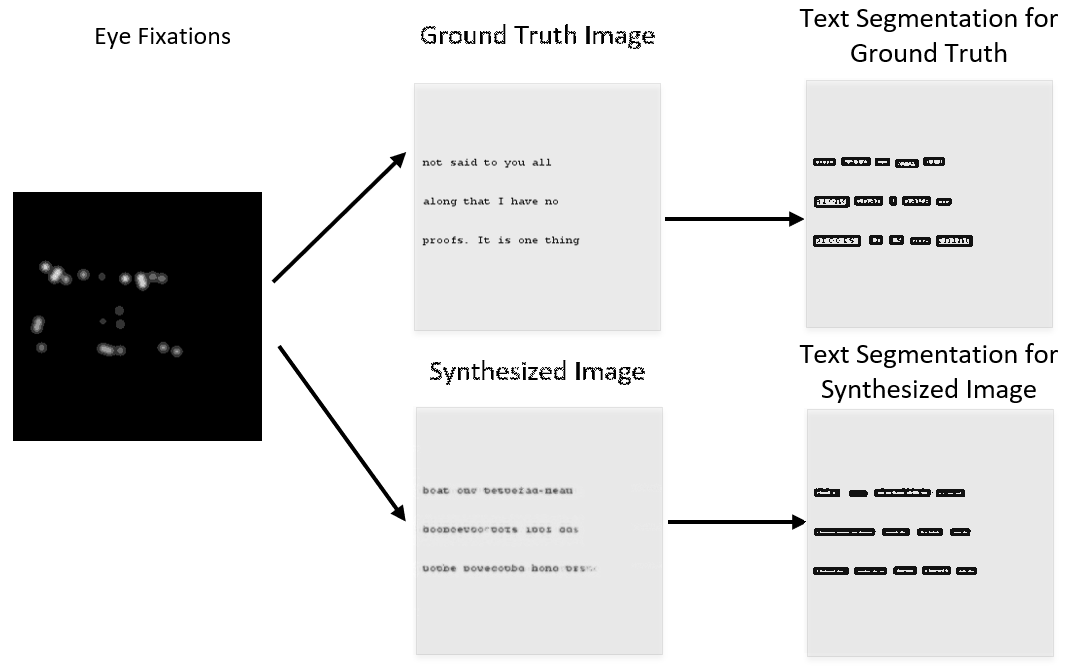}}
\end{center}
   \caption{Illustration of synthesized versus ground truth text-embedded images, along with the segmentation process used to obtain the word segment length distribution.}
\label{fig:graph_photo_for_seg_length}
\end{figure}

As for evaluating the performance of our approach when it is applied to the GECO dataset, we first provide some example results of the generated text-embedded images in Figure~\ref{fig:exampleresults}, from which a qualitative evaluation can be done. It is important to remember that, although the network is designed to perform image generation without using any natural language processing, the training has proven to be extremely successful in the sense that all the generated images display text-like areas. Furthermore, numerous valid English words can be observed in the generated images. We also utilized various text analysis metrics, in combination with human inspection, in order to evaluate the quality of our outputs. Optical Character Recognition (OCR) was used to extract all the text-like content from the synthetic images generated by our network \cite{smith2007overview}. In total, 13270 alphabet characters are retrieved from 241 synthetic images. There are only 72 occurrences for which OCR engine is not able to convert a scanned character into a valid alphabet letter. In addition, as presented in Figure~\ref{fig:graph_photo_for_seg_length}, we also used text segmentation in attempt to divide the text content in our synthesized images into segments with various lengths. Each segment can be considered as an individual word that the system generates based on the input eye fixation data. 

It is worth mentioning that for illustration purpose, we display the qualitative results of the text-embedded images by using a black text color with white background color. Although we used a red and green encoding method to facilitate the training of our network, the generated results can be easily converted to black and white afterwards in order to be more similar to the real image content that is viewed by the readers. 

Figure~\ref{fig:combo_histo} compares the histogram that summarizes the length distribution of all segments obtained from our synthetic images versus the histogram generated using the ground truth images. It can be observed that the shape of the two distributions are similar. Most segments have lengths ranging from 23 to 26 pixels. This means that in both ground truth and generated images, most words are constructed using 3 to 4 characters. Thus we observe a close similarity between the generated images and the ground truth images. This serves as a demonstration of our network's ability to generate highly text-like image content.

\section{Conclusion}

In this paper, we explored the possibility of inverting the relationships between image stimuli and eye movements and successfully developed an approach to synthesize the content of viewed images based on eye tracking information. We presented WAYLA, a deep-learning technique which utilizes Conditional Generative Adversarial Nets to generate newspaper images with different detail levels from eye fixation heat maps. Moreover, the system is able to reconstruct images containing only text using solely eye tracking data. Our work proved the feasibility of generating image content from gaze data, which has never been demonstrated before. Our results showed that the deep convolutional neural network that we used is effective at performing image generation tasks, achieving high similarity between the generated results and the ground truth. Although further improvements could be achieved to enhance the quality of the generated images, the idea of inverting the path starting from viewed content and ending with eye tracking information can be widely applied in various settings. For example, many previous works use Baysian Inference \cite{carpenter1991visual} as the standard approach to infer visual task from eye movements. The proposed WAYLA approach can be used to improve the performance of these traditional methods. Furthermore, previously most studies related to image generation using Generative Adversarial Networks and its derivatives were done without considering gaze information as input. By introducing the possibility of inferring image content based on gaze data, the WAYLA approach could open doors for more diversified image generation models in the future. 

{\small
\bibliographystyle{ieee}
\bibliography{egbib}
}

\end{document}